\documentclass[conference]{IEEEtran}
\usepackage{graphicx}
\usepackage{graphicx}
\usepackage{amsmath}
\usepackage[amsmath]{ntheorem}
\usepackage{lastpage}
\usepackage{amsfonts}

\usepackage{amssymb}
\usepackage{graphicx}
\usepackage{hyperref}
\usepackage{amsmath}
\usepackage[amsmath]{ntheorem}
\usepackage{lastpage}
\usepackage{amsfonts}
\usepackage{amssymb}
\usepackage{graphicx}
\usepackage{hyperref}
\usepackage{amssymb}
\usepackage{lineno}
\usepackage{amsmath}
\usepackage[utf8]{inputenc}
\usepackage{fixltx2e}
\usepackage{graphicx}
\usepackage{tabularx}
\usepackage{subcaption}
\usepackage{graphicx}
\usepackage[utf8]{inputenc}
\usepackage{hyperref}
\numberwithin{equation}{section}
\usepackage{lipsum}
\usepackage{hyperref} 
\makeatletter
\usepackage{caption}
\usepackage{indentfirst}
\newcommand{\mathleft}{\@fleqntrue\@mathmargin0pt}
\newcommand{\mathcenter}{\@fleqnfalse}
\usepackage{lineno,hyperref}
\modulolinenumbers[5]
\usepackage{multirow}
\usepackage[table,xcdraw]{xcolor}
\usepackage{colortbl}
\usepackage{tabularx}
\usepackage{rotating}
\usepackage{graphicx}
\usepackage{multirow}
\mathleft

\ifCLASSINFOpdf
\else
\fi
\begin{document}

\title{Sentimental Content Analysis and Knowledge Extraction from News Articles}

\author{\IEEEauthorblockN{Mohammad Kamel}
\IEEEauthorblockA{Center of Excellence on Soft Computing\\ and Intelligent Information Processing\\Ferdowsi University of Mashhad, Iran\\
Email: mohammad.kamel@mail.um.ac.ir}
\and
\IEEEauthorblockN{Neda Keyvani}
\IEEEauthorblockA{Department of Mathematical Science\\ Ferdowsi University of Mashhad, Iran\\
Email: neda.keyvani@mail.um.ac.ir}
\and
\IEEEauthorblockN{Hadi Sadoghi-Yazi}
\IEEEauthorblockA{Center of Excellence on Soft Computing\\ and Intelligent Information Processing\\Ferdowsi University of Mashhad, Iran\\
Email: h-sadoghi@um.ac.ir}}

\maketitle

\begin{abstract}
In web era, since technology has revolutionized mankind life, plenty of data and information are published on the Internet each day. For instance, news agencies publish news on their websites all over the world. These raw data could be an important resource for knowledge extraction. These shared data contain emotions (i.e., positive, neutral or negative) toward various topics; therefore, sentimental content extraction could be a beneficial task in many aspects. Extracting the sentiment of news illustrates highly valuable information about the events over a period of time, the viewpoint of a media or news agency to these events. In this paper an attempt is made to propose an approach for news analysis and extracting useful knowledge from them. Firstly, we attempt to extract a noise robust sentiment of news documents; therefore, the news associated to six countries: United State, United Kingdom, Germany, Canada, France and Australia in 5 different news categories: Politics, Sports, Business, Entertainment and Technology are downloaded. In this paper we compare the condition of different countries in each 5 news topics based on the extracted sentiments and emotional contents in news documents. Moreover, we propose an approach to reduce the bulky news data to extract the hottest topics and news titles as a knowledge. Eventually, we generate a word model to map each word to a fixed-size vector by Word2Vec in order to understand the relations between words in our collected news database.
\end{abstract}


\IEEEpeerreviewmaketitle

\section{Introduction}
\par There are a number of news agencies throughout the world which publish the international and domestic news each day. People in recent years have understood that even the events and news of other countries may have direct effect on their lives; therefore, they track a wide range of news sources. In web era, the researches reveal that Internet as a new platform for sharing news is more popular than newspapers; hence, understanding all the shared information and extracting the knowledge contained in the Web or social networks especially online news agencies could be beneficial in many ways. To this end, natural language processing (NLP) could be utilized to analyze and extract knowledge form text documents. 
\par Natural language processing is a field that covers computer understanding and manipulating of human language. Entity extraction, part of speech tagging, sentiment analysis, and word modeling are amongst the most famous NLP algorithms. The sentiment analysis tasks are defined as extracting the mood and emotion of textual data written by a writer. Sentiment analysis techniques are divided into two clusters, learning based and lexical based approaches. Firstly, lexical based sentiment analysis tools use a set of words and phrases with positive or negative numeric score. In lexical based methods, sentiments are analyzed by using the frequencies of positive and negative words in a text[1-3]. In the second group which are learning based approaches, an attempt is made to apply machine learning algorithms such as support vector machines, neural networks and etc. [4-7].
\par The task of sentiment analysis can be divided into some subdivisions as term extraction, category detection, sentiment classification and sentiment rating. The purpose of aspect rating which is in the main focus of this paper is to assign a numeric rating (i.e. 1 $\sim$ 5 stars or in our research i.e. -2 $\sim$ 2 stars) which -2 is the most negative sentiment and +2 is the most positive sentiment[8], [9]. These researches give birth to creation of many tools that can be used for sentiment analysis such as CoreNLP, textblob, Spacy and natural language toolkit (NLTK).
\par Since the data published by news agencies or social networks are bulky, instance selection is a necessity to reduce the amount of data and prepare the datasets and textual data to a manageable volume for extracting information and knowledge. As defined, instance selection is an important data preprocessing step which could be applied for reducing original datasets to a manageable volume and removing noisy instances, before applying learning algorithms [10]. As is self evident, instance selection approaches could be divided into three different groups: 1- algorithms which attempt to eliminate noisy data, 2- algorithms which attempt to select and manipulate effective data from original dataset and 3- clustering based algorithms. It worth to mention that instance selection algorithms are mainly based on k nearest neighbor (KNN). For instance, condensed nearest neighbor (CNN), reduced nearest neighbor algorithm (RNN), selected nearest neighbor (SNN), generalized condensed nearest neighbor (GCNN) and edited nearest neighbor (ENN) are amongst the most well known KNN based instance selection approaches[11-16].
\par In addition, one of the main NLP tasks is document classification [17]. For this purpose, we need to extract appropriate features from each document and employ them as the input of a classifier. There are several effective machine learning techniques for data classification for example Support Vector Machines (SVM) [18]. Moreover, feature extraction is a challenging part, since each document can be processed in different levels of abstraction and from syntax and semantic points of view. Word2Vec [19] is an effective method to map each word to a fixed-size numeric vector. Combining the corresponding vectors of main vocabularies of a document could be a suitable feature of that document.
\par The main purpose of this paper is to propose a tool for sentimental content analysis of the world news and extracting knowledge from them. Firstly, the focus is on extracting the sentimental content of the news documents to get aware of politics, business, entertainment, sports and technology conditions in different countries considering the associated news to them. The selected countries are United State, United Kingdom, Germany, Canada, France and Australia. The approach could be beneficial for experts to compare and contrast the conditions existing in their country with other countries for better planning and decision making. To this end we collected the news related to six countries by using webhose API; afterward, CoreNLP, a state-of-the-are tools in natural language processing, is utilized to extract the sentimental content of the news. Since the sentiment analysis tools mostly extract sentiment for sentences separately with out any attention to the context of the document, we employed some fusion methods to a have a value as the whole document sentiment; meanwhile, these fusion methods attempt to remove the noise of both sentiment analysis tools and writers (noise of writers means off topic sentences).
\par Secondly, in this paper an attempt is made to reduce bulk of news data in order to select the hottest news topics in six selected countries. Moreover, We also generate a Word2Vec model based on the collected news in order to understand the words relationship.
\par The rest of this paper is organized as follows: the techniques of collecting the data is presented in Section 2. The methodology of sentimental content analysis and the results are introduced in Section 3. Section 4 presents the methodology and results of data reduction step on the news data. Moreover, section 5 introduces the technique of word modeling. Eventually, conclusion of the paper is presented in section 6.
\section{Data gathering from news websites}
\par In this section an attempt is made to introduce the method of data gathering, data cleaning and data preprocessing. For data gathering phase, a python crawler called webhose\footnote [1] {https://docs.webhose.io} is used. The webhose API provides access to structured web data feeds across vertical content domains. This API offers multiple data repositories such as news, blogs, discussions, e-commerce and dark web content. By using the mentioned API, we are capable to download news from different countries with different news categories in  many languages; furthermore, the researcher can adjust the time interval to download the news considering his/her need. For example an one month time interval is used in this paper. It worth to mention that an advantage of this API is the variety of data from different news agencies with different attitudes towards topics which makes the results of the research more reliable. Table 1 reveals the number of collected news in English language from six selected countries consist of United State, United Kingdom, Germany, Canada, France and Australia in 5 different news categories consist of politics, sports, business, entertainment and technology. The next step is data cleaning, which consists of normalizing the textual data, such as lemmatization and stemming. Eventually, in the preprocessing step, the consequent steps which are mentioned below are performed: 1) removing sentences that linguistically are not in English; 2) remove nonascii symbols; 3) eliminating the sentences which their Length are less than 5 words. Figure 1 demonstrates the main steps of data gathering phase.
\begin{center}
	\small\addtolength{\tabcolsep}{-3pt}
	\begin{table}[h!]
		\caption{The number of downloaded news in English language from six selected countries in 5 different news categories in an one month time interval.}
		\begin{tabular}{ | c | c | c | c | c | c | c| }
			\hline
			Country/Category & Politics & Sports & Technology & Entertainment & Business \\
			\hline \hline
			United State & 770110& 200659 & 393621 & 254264 & 194293 \\ 
			United Kingdom & 28333 & 137950 & 35093 & 40067 & 46092  \\ 
			Germany & 9176 & 11722 & 18003 & 7983 & 21404 \\ 
			Canada & 1368 & 4285 & 5276 & 8200 & 13114\\
			France & 1097 & 12461 & 5159 & 9231 & 3742 \\
			Australia & 1975 & 17574 & 3245 & 15776 & 110876 \\
			\hline

		\end{tabular}
		
	\end{table}
\end{center}
\begin{figure}[h!]
	\includegraphics[width=5.5cm ,height=9.5cm]{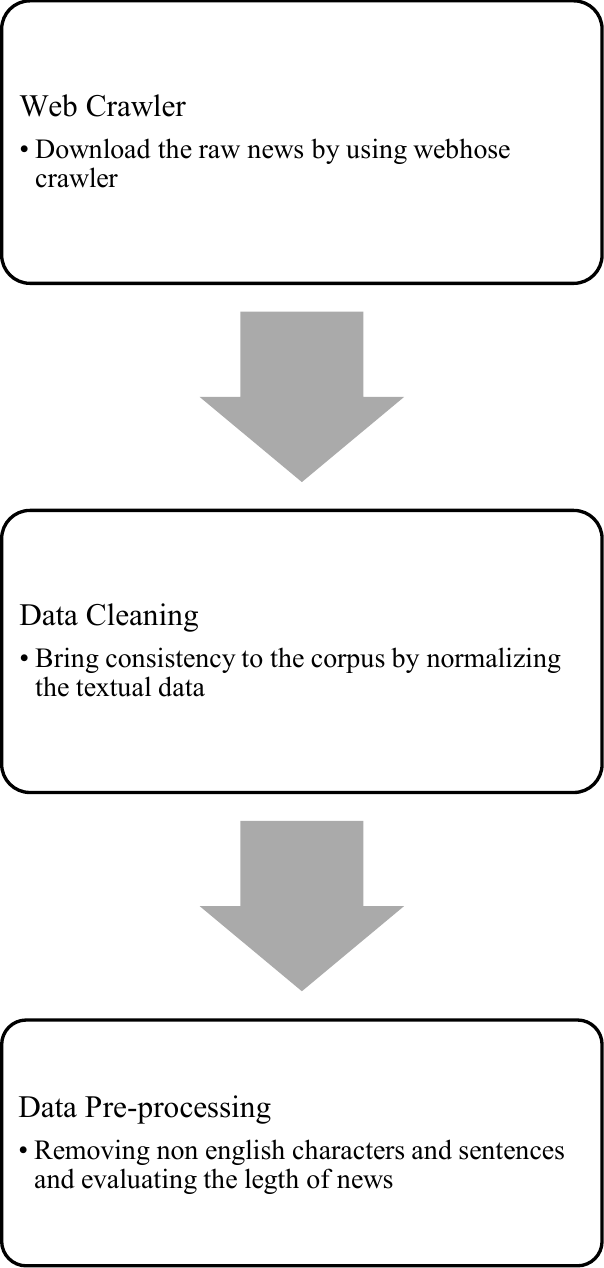}
	\centering
	\label{fig:1}
	\caption{The main steps of data gathering phase. }
\end{figure}

\section{sentimental content extraction}
\par The purpose of this section is to extract and analyze the sentiment of shared news in six selected countries. The results are beneficial for society analysts and sociologists to be aware of politic, entertainment, sport, technology and business conditions in different countries by using the news which are shared in social networks or news websites. Moreover, politicians could utilize the proposed method of this paper to compare and contrast the politic, entertainment, sport, technology and business of their own country with others. Emotional contect analysis could guide experts for better planning or decision making. This section consists of three subdivisions: 1) sentiment analysis, 2) sentiment fusion and 3) sentimental content analysis.
\subsection{Sentiment Analysis}
\par There are several toolboxes, API's and Techniques which could be used to analyze the
sentiment of texts. In this paper CoreNLP, one of the most novel and efficient methods
for sentiment analysis, is utilized. It worth to mention that CoreNLP is a learning based approach in sentiment analysis and text processing. CoreNLP was established based on recursive neural tensor networks and the Stanford Sentiment treebank which includes 215,154 phrases extracted from 11,855 sentences[20]. Figure 2 demonstrates a demo for analyzing the sentiment of sentences\footnote [2] {http://nlp.stanford.edu/sentiment/treebank.html}. The results of CoreNLP for each sentence is between -2 to +2 which -2, -1, 0, 1, 2 are very negative, negative, neutral, positive and very positive sequentially.
\begin{figure}[h!]
	\includegraphics[width=9cm ,height=5cm]{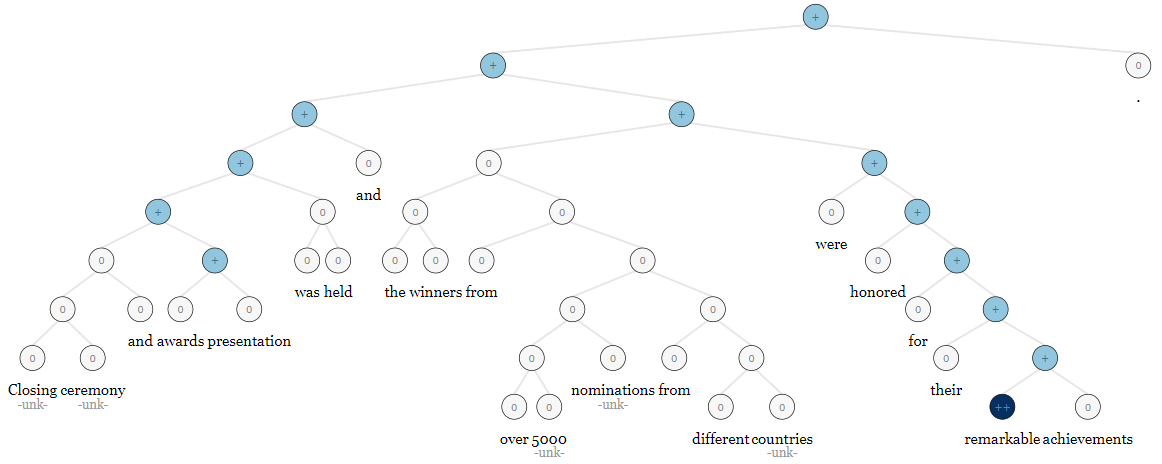}
	\centering
	\label{fig:2}
	\caption{Parse tree and sentiment analysis of a sample sentence using Stanford sentiment analysis tool.}
\end{figure}
\subsection{Sentiment Fusion}
\par As is self evident, most of the sentiment analysis tools are sentence based, in another words, they extract the sentiment for sentences separately with out any attention to the context of the text. It is obvious that most of the time each news text consists of some sentences and paragraphs; moreover,  in extracting the sentiment of a news text, considering the sentiment of all sentences is a necessity. After extracting the sentiment of sentences in a news with sentiment analysis tools, in this paper CoreNLP is utilized, we need an approach to fuse all separated sentiments to have a value as the whole text sentiment. We have to use an approach that not only fuse the sentiment of separated sentences to have a final value as the document sentiment but also remove the possible noises. Initial experiments have shown that the source of these noises could be the sentiment analysis tools or writers while writing off topic sentences [21].
\par In this paper we used average operator and correntropy loss to fuse all separated sentiments value to have a sentiment for the whole document [21-22]. It worth to mention that by using correntropy loss, we can diminish the effects of noisy signals to have a noise robust sentiment for the whole news document. The fusion formulas could be written as follows.
\\\\
\textbf{Average Operator}:
\\
\begin{equation}	
\mu = \frac{\sum_{i=1}^{N} x\textsubscript{i}}{N}
\end{equation}
\\
\textbf{Correntropy Loss}:
\\
\begin{equation}
p_i = - exp(-\eta(x_i-\mu)^2)
\end{equation}
\begin{equation}
\mu = \frac{\sum x_i p_i}{\sum p_i}
\end{equation}
\\
\par Where $N$ is the number of sentences in a news and $x_i$ is the extracted sentiments of sentences in a news document. In addition, Figure 3 demonstrates the extracted sentiments of a selected news and the final values as the whole news sentiment based on different loss functions. As could be seen some of the extracted sentiments are positive while the majority are negative. It could be concluded that the news totally is about a negative event and positive sentiments might be noises due to the noise of sentiment analysis tools or off topic sentences; therefore, the effects of them in the final document sentiment have to be diminished and it is what correntropy loss have done.
\begin{figure}[h!]
	\includegraphics[width=7cm ,height=4cm]{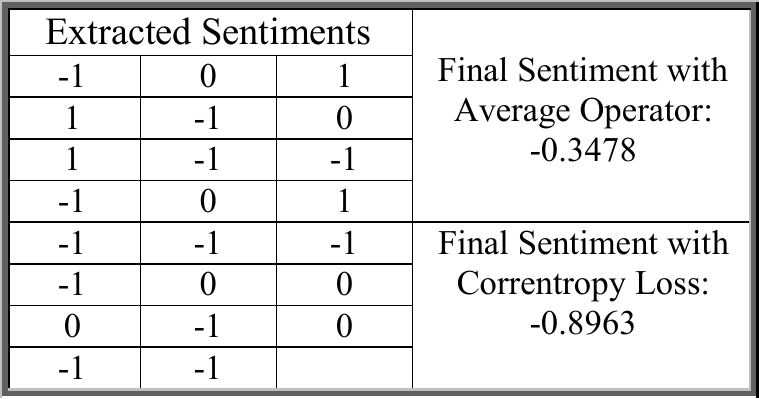}
	\centering
	\label{fig:3}
	\caption{The extracted sentiments of a selected news and the final values as the whole news sentiment based on different loss functions.}
\end{figure}
\\
\subsection{Sentimental Content Analysis}
\par Our collected data in this paper are English news in five different categories in the interval of one month from six selected countries. It is valuable to mention that the researcher who use the webhose API can adjust the time interval for data gathering phase considering his/her need. The collected news are published in social networks or news agencies website of the studied countries. It worth to mention that the ultimate goal of this paper is to propose an approach for automatic analyzing the associated news to the six selected countries and extracting knowledge from them. To this end the webhose API is utilized. A sample query to download news from webhose API is demonstrated in Figure 4.
\begin{figure}[h!]
	\includegraphics[width=9cm ,height=1.5cm]{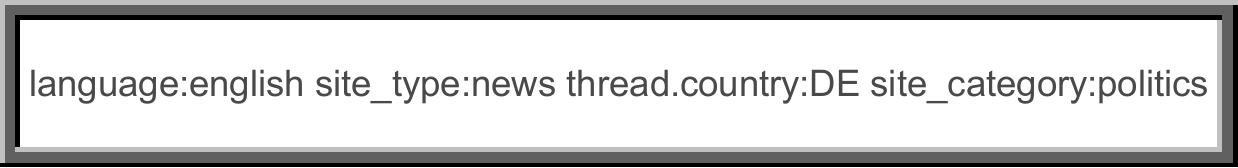}
	\centering
	\label{fig:3}
	\caption{A sample query to download news from webhose.io website. In this sample query we set some conditions to be taken into consideration while downloading news. The query will download English news which their category is politics and are associated to Germany in the adjusted interval.}
\end{figure}
\par After data gathering phase, we employ NLTK as a general-purpose sentiment analysis tools to extract the sentiments of separated sentences in a news. Afterward, by using some fusion methods which were described previously we extract a sentiment as the sentiment of a news document. Eventually, in the final step we calculate a final sentiment of news for different categories in six selected countries. It worth to mention that the The determined sentiments are scaled to the interval of zero and one. The main steps of emotional content extraction are demonstrated in Figure 5. Figure 6 represents the noise robust final sentiments of news in different categories amongst six selected countries. As an example, it could be concluded from Figure 6 that in United State the news related to sports are the most positive news; while, politics news have the most negative sentiment. The ranking of the published news which are related to United State from the most positive to the most negative news category based on our analysis is sports, technology, business, entertainment and politics respectively. As an another example in Canada technology, entertainment, business, politics and sports have the most positive emotional content to the most negative, sequentially.
\begin{figure}[h!]
	\includegraphics[width=8.75cm ,height=3.5cm]{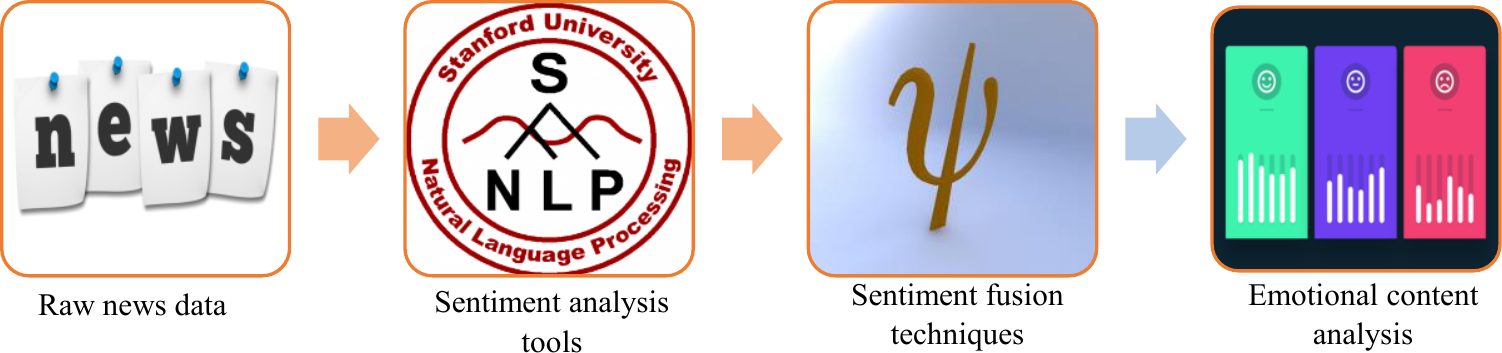}
	\centering
	\label{fig:3}
	\caption{Flowchart of the process: Different steps are shown. Refer to text for more details.}
\end{figure}

 \par Figures 7 is a comparison chart between countries and different categories. For instance, the best country in entertainment news is Australia; while, Germany has the most negative entertainment news. As an another example, United state has the best business condition amongst the studied countries. The France, Germany, United Kingdom, Canada and Australia are in the next business rank due to the emotional content of the associated news to them. To sum it up, it could be concluded that United State is the most positive country in sports and business, the Australia is most positive country in entertainment, the Canada is most positive country in technology and the Germany is the most positive country in politics considering the extracted emotional content from the associated news to them in the selected time interval.
\begin{figure}[h!]
	\includegraphics[width=9cm ,height=7cm]{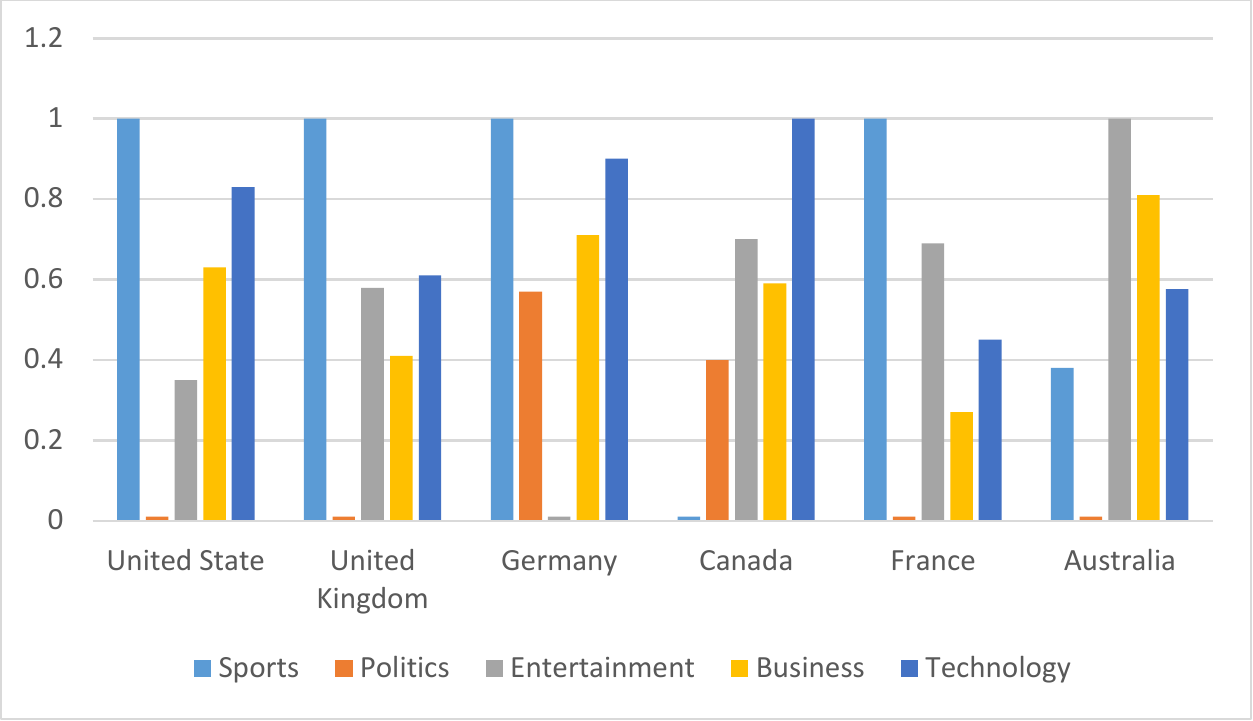}
	\centering
	\label{fig:3}
	\caption{The sentiments of news based on different categories in six selected countries. The extracted sentiments for each country are scaled to the interval of zero and one separately.}
\end{figure}

\begin{figure}[h!]
	\includegraphics[width=9cm ,height=7cm]{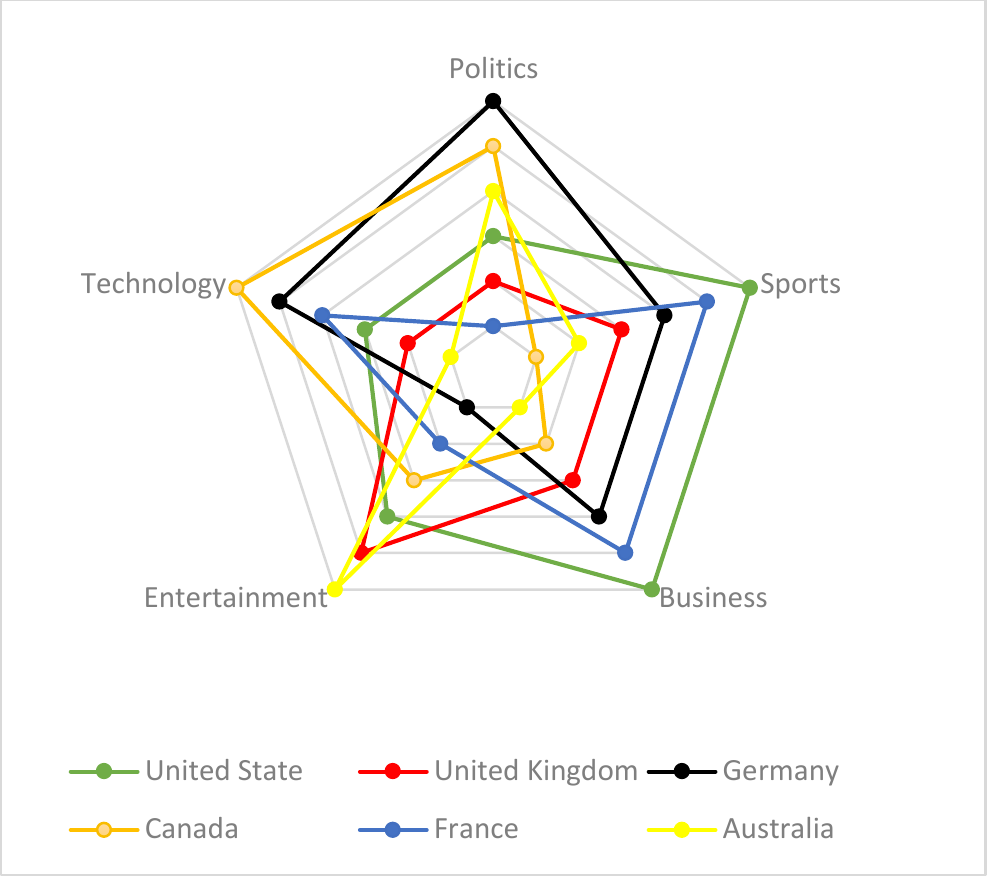}
	\centering
	\label{fig:3}
	\caption{Represent the percentages of different emotional contents in different countries using the radar	chart. There are a sequence of equiangular spokes, called radii.}
\end{figure}

\section{data reduction}
\par The purpose of this section is to reduce bulk of news data in order to select the hottest news topics in six selected countries. The results of this section could be beneficial for political analysts, politicians, sociologists and other major professions to be aware of different countries condition in different fields related to their majors. To extract such important knowledge and information, an attempt is made to use data reduction algorithms.
\par In this paper we utilized Decremental Reduction Optimization Procedure 3 (DROP3) algorithm as one of the most efficient data reduction algorithm [23]. In this algorithm, each instance $x$ has $k$ nearest neighbors; moreover, the instances that have x as one of their $k$ nearest neighbors are called the associates of $x$. Initially, DROP3 algorithm executes ENN to filter noises. Afterward, the DROP3 algorithm removes an instance $x$ under some conditions.
\par To reduce textual data to a manageable volume; initially, the feature vector of textual data have to be extracted. By this means different algorithms such as word2vec, tf-idf or etc. could be utilized. In statistics tf-idf is an approach that is performed to reflect how important a word is in a document or corpus. The tf-idf value is dependent to how many times a word is repeated in a document. Furthermore, it helps to adjust for the fact that some words appear more frequently and it is used as a prominent term-weighting schemes in text mining. The formula of tf-idf could be written as follows:
\\
\begin{equation}
tf(t,d)= f_{t,d}/\Sigma f_{t\prime,d}
\end{equation}
\begin{equation}
idf(t,D)=\log({N/n_t})
\end{equation}
\begin{equation}
tfidf(t,d,D)=tf(t,d) . idf(t,D)
\end{equation}
\\
Where $f_{t,d}$ is the number of times term t appears in the document d, $f_{t\prime,d}$ is the total number of terms in the document d, $N$ is the total number of documents and $n_t$ is the number of documents with term t in it.
\par In this paper we used tf-idf to convert title of news into feature vectors which now could be used as an input for data reduction step. After extracting the feature vectors of news of different countries in five different category, we can perform DROP3 approach to reduce our data. As mentioned previously the output of this section are the hottest news title which are selected by data reduction algorithms. It is valuable to mention that the proposed approach could be performed on different parts of a news such as news body. Figure 8 demonstrates the selected title of the published news in the studied interval. 

\begin{figure}[h!]
	\includegraphics[width=9cm ,height=12cm]{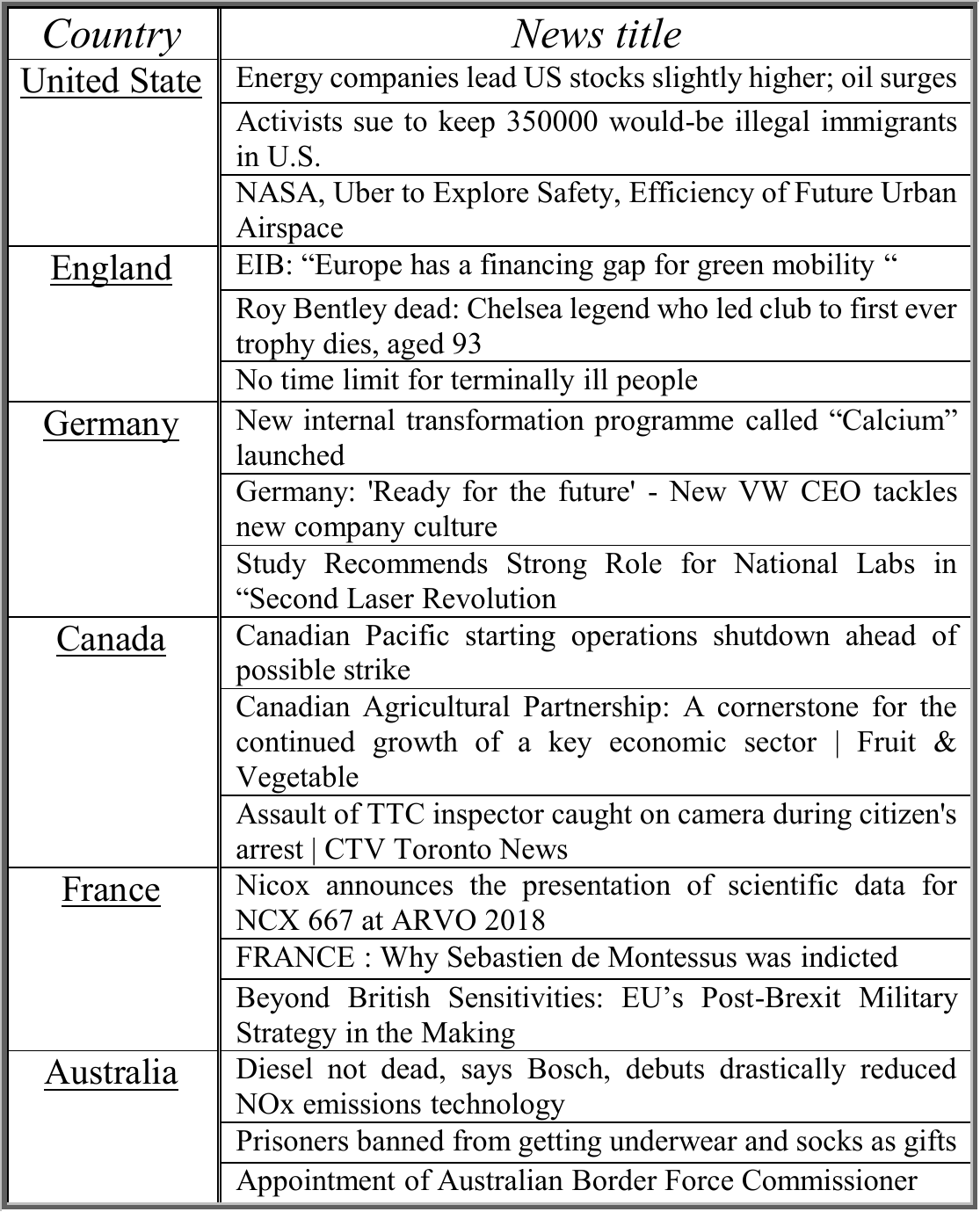}
	\centering
	\label{fig:3}
	\caption{Selected news title from collected news after performing instance selection algorithm.}
\end{figure}

\section{word modeling}
\par Mapping words to a fixed-size numeric vector is a useful representation for text classification and knowledge visualization. In this section, we aim to map each word of the collected news data to a numeric vector. For text classification, the average of corresponding vectors of each word of a document can be an appropriate representation. Moreover, by mapping this vector representation of words to a 2-D space, we can visualize words in order to understand their relations better. One of the state-of-the-art models for this purpose is Word2Vec. This model can map each word of the dictionary to a numeric vector. Using 50 dimension in the latent space, our trained network is capable of creating a suitable feature space that the closer words have related. Table II illustrates the results of a sample query for three words. According to this table, our model can effectively cluster the words.
The main steps of knowledge extraction phase is demonstrated in Figure 9.
\begin{table}[h!]
	\begin{center}
		\small\addtolength{\tabcolsep}{+2pt}
		\caption{The closest word to three sample queries.}
		\label{tab:table1}
		\begin{tabular}{c|c|c} 
			\textbf{United State} & \textbf{world} & \textbf{peace} \\
			\hline
			\hline
			\hline
			\\
			power & global & commitment\\
			growth & nature & diplomatic\\
			p5+1 & relation & tranquility\\
			force & population  & harmony\\
			negotiation & grouping & confederation\\
		\end{tabular}
	\end{center}
\end{table}

\begin{figure}[h!]
	\includegraphics[width=8cm ,height=4cm]{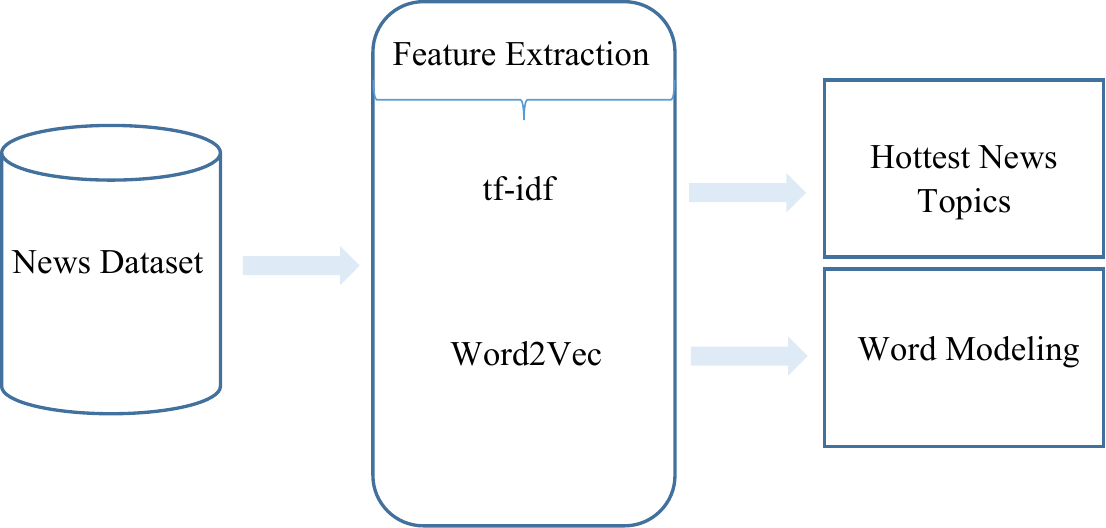}
	\centering
	\label{fig:3}
	\caption{The main steps of knowledge extraction phase.}
\end{figure}


\section{conclusion}
\par The purpose of this paper is to propose an approach for news analysis and extracting useful knowledge from them. Initially, we employed natural language processing techniques and some fusion methods to extract the sentiment of news in order to make an emotional ranking of countries with respect to the news associated to them. Moreover, we proposed an approach to reduce news data to select the hottest news topics based on instance selection algorithms. Eventually, we create a word model using Word2Vec for our collected news corpus. The results and analysis of this paper could be beneficial for society analysts, sociologists and politicians to extract knowledge about politic, entertainment, sport, technology and business conditions throughout the world considering the news articles.

\end{document}